%% file: root.tex
\documentclass[letterpaper, 10 pt, conference]{ieeeconf}  

\IEEEoverridecommandlockouts                              

\usepackage{amsmath} 
\usepackage{amssymb}  
\usepackage{subcaption}
\usepackage[noend]{algpseudocode}
\usepackage{algorithm}
\usepackage{booktabs}
\usepackage{multirow}
\usepackage{color}

\makeatletter
\newcommand\fs@betterruled{%
  \let\@fs@capt\floatc@ruled
  \def\@fs@pre{\vspace*{5pt}\hrule height.8pt depth0pt \kern2pt}%
  \def\@fs@post{\kern2pt\hrule\relax}%
  \def\@fs@mid{\kern2pt\hrule\kern2pt}%
  \let\@fs@iftopcapt\iftrue}
\floatstyle{betterruled}
\restylefloat{algorithm}
\makeatother

\title{\LARGE \bf
Knowing When to Ask: Resolving Uncertainty in Human-Robot Joint Planning via Explicit Dialogue and Implicit Intent Cues
}

\author{Zeyu Fang$^{1}$, Yuxin Lin$^{3}$, Cheng Liu$^{2}$, Beomyeol Yu$^{2}$, Zeyuan Yang$^{2}$, Rongqian Chen$^{1}$, \\ Taeyoung Lee$^{2}$, Mahdi Imani$^{3}$ and Tian Lan$^{1}$%
\thanks{$^{1}$Zeyu Fang, Rongqian Chen, and Dr. Tian Lan are with Department of Electrical and Computer Engineering, George Washington University
       {\tt\small joey.fang@gwu.edu, rongqianc@gwu.edu, tlan@gwu.edu}} 
       \thanks{$^{2}$Cheng Liu, Beomyeol Yu, Zeyuan Yang and Dr. Taeyoung Lee are with the Department of Mechanical and Aerospace Engineering, George Washington University
       {\tt\small chengl@gwu.edu, yubeomyeol@gwu.edu, zeyuan.yang@gwu.edu, tylee@gwu.edu}}%
\thanks{$^{3}$Yuxin Lin and Dr. Mahdi Imani are with the Department of Electrical and Computer Engineering, Northeastern University
       {\tt\small lin.yuxi@northeastern.edu, m.imani@northeastern.edu}}%
 }

\usepackage{graphicx}

\begin{document}

\maketitle

\thispagestyle{empty}
\pagestyle{empty}

\begin{abstract}

Effective human-robot collaboration in open-world environments requires joint planning under uncertainty about the task, the environment, and the human teammate. Communication is the most direct means of resolving such uncertainty, yet most existing systems support only one-way communication: robots listen and act, treating humans as passive supervisors rather than conversational teammates capable of two-way dialogue. We propose a unified human-robot joint planning system in which the robot actively resolves uncertainty through two complementary communication channels. When uncertainty is decision-critical, an uncertainty-mitigation joint planning module engages the human in clarification dialogue: it grounds ambiguous instructions via an LLM-assisted active elicitation mechanism, enumerates traversability hypotheses through a hypothesis-augmented $A^*$ search, and computes a cost-optimal querying policy via dynamic programming, so that the robot asks only the questions whose answers actually matter for the plan. When explicit dialogue is unnecessary or impractical, a real-time intent-aware collaboration module instead reads implicit, nonverbal cues, maintaining a probabilistic belief over the human's latent task intent from spatial and directional signals to enable coordination-aware task selection without any communication overhead. We validate the proposed system in both Gazebo simulations and real-world UAV deployments, integrated with a voice dialogue interface and a Vision-Language Model (VLM)-based 3D semantic perception pipeline. Experimental results show that cost-optimal clarification dialogue cuts the interaction cost by $51.9\%$ while maintaining a $100\%$ task success rate, and implicit intent reading reduces the cooperative task execution time by $25.4\%$ compared to the baselines.

\end{abstract}


\input{introduction}

\input{methodology}

\input{experiments}

\section{Conclusion}

In this paper, we propose an end-to-end, unified human-robot joint planning system that resolves the dual sources of uncertainty in open-world collaboration through two complementary communication channels.
By knowing when to ask and when to stay silent, our system enables autonomous agents to act as proactive, conversational teammates rather than passive tools.
Within the core planning engine, the uncertainty-mitigation joint planning module resolves decision-critical uncertainty through explicit clarification dialogue with strictly minimized interaction costs, while the real-time intent-aware collaboration module infers the human's latent objectives from implicit spatial and directional cues, continuously updating a probabilistic belief for adaptive task coordination without communication overhead.
Extensive evaluations in both Gazebo simulations and real-world UAV deployments show that our system significantly outperforms conventional baselines in both modes, reducing the required interaction cost by 51.9\% and accelerating the total cooperative task execution time by 25.4\%. 

Future work will focus on tightly coupling these two modes, allowing the robot to dynamically transition from implicit intent inference to explicit querying when behavioral confidence drops, and scaling the framework to multi-agent, multi-human scenarios under complex temporal constraints.




\bibliographystyle{ieeetr}
\bibliography{references}

\end{document}

%% file: introduction.tex
\section{Introduction}


\begin{figure*}[t]
    \centering
    \vspace{0.1cm}
    \includegraphics[width=\textwidth]{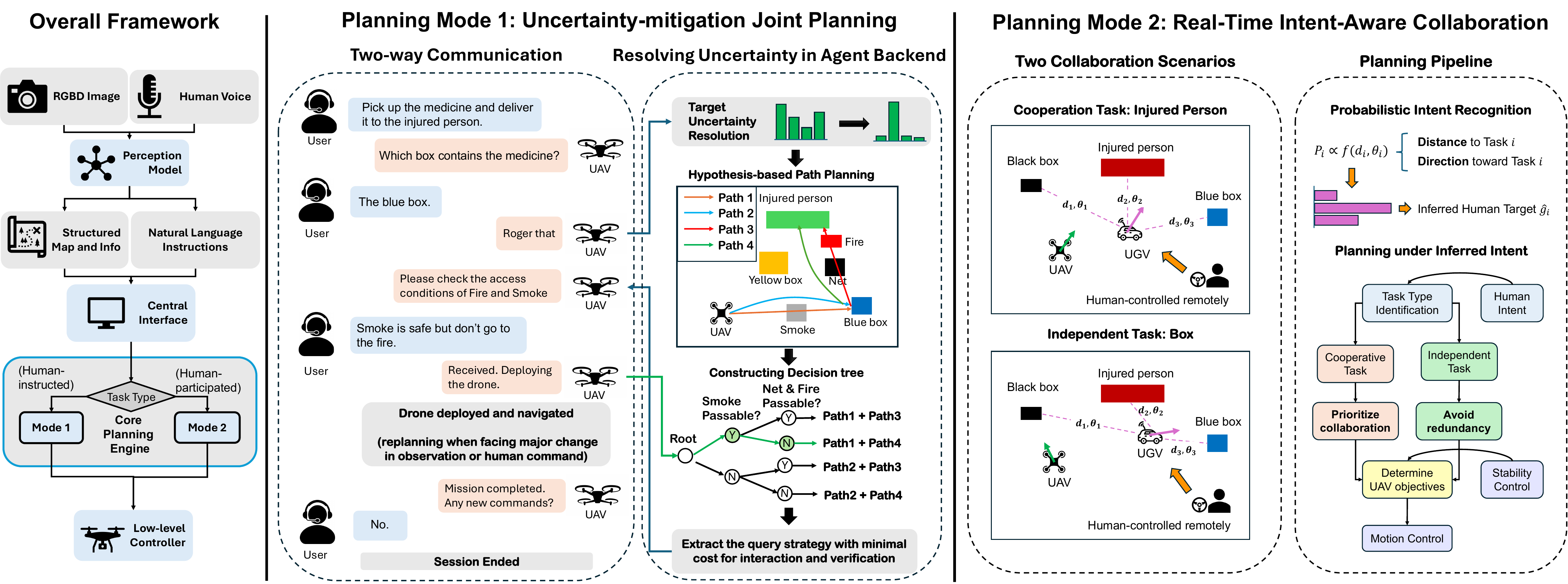}
    \caption{Overview of the proposed system and the two planning modes inside the core planning engine. The system first analyzes the surrounding environment via perception models; the central interface then activates one planning mode based on the task type. The planning results, either way-point paths or high-level goals, are sent to the low-level controller to drive the drone. Details of the two modes are elaborated in Section \ref{chapter:Method}.
    }
    \label{fig:overview}
    \vspace{-0.6cm}
\end{figure*}







Human–robot collaboration has become a fundamental capability for autonomous systems in real-world missions such as healthcare, manufacturing, and disaster response~\cite{berretta2023defining}.
AI agents are expected to function as human-like teammates that are capable of communication, understanding, and cooperation, instead of simply being tools that amplify human capabilities or generate human-like responses~\cite{fang2024coordinate}.
However, open-world human-robot joint planning often faces significant challenges from uncertainties: task-relevant factors (e.g., object properties, instruction semantics, and human intent) are often ambiguous, obscure, or partially observable, creating knowledge gaps that can alter transitions, constraints, and goals within the deployed environment, and thus degrade planning quality if ignored or handled improperly~\cite{aghzal2025survey}.

A robot that can only listen and act is poorly equipped to close these knowledge gaps: it may misground an ambiguous instruction, plan over false assumptions, or duplicate its teammate's effort. Asking targeted clarification questions resolves such uncertainty before the robot commits to a plan. Yet dialogue is not free: each query consumes human attention and interrupts execution, so an effective teammate must know \emph{when to ask, what to ask, and when to stay silent}. Nor is spoken language the only channel: human motion itself carries rich intent information that a robot can read without imposing any interaction cost. We therefore treat explicit clarification dialogue and implicit cue reading as two complementary channels on a single communication spectrum, and design a joint planning system that exploits both.

Categorized by the modality of human involvement, existing methods on human-robot collaboration can be broadly bifurcated into two approaches: explicit human-guided planning, where human input is restricted to high-level instructions, and implicit human intent inference, where the human directly participates in the task execution, and the robot infers human intent only from its observations without communication~\cite{obaigbena2024ai, li2026acdzero}.
For the human-guided approach, traditional methods to incorporate human feedback into planning, including Human-In-The-Loop (HITL)~\cite{mosqueira2023human} and human-aware task planning~\cite{lemaignan2017artificial}, largely treat humans as passive supervisors providing expert demonstrations or corrections, instead of facilitating task-relevant knowledge exchange through two-way communication~\cite{huang2022language}.
Recent work leverages large language models (LLMs) to translate natural-language instructions into executable robot actions~\cite{fang2026mint}, extends static goal prediction to continual inference under evolving human preferences~\cite{ghose2025ve}, and frames joint planning as a bidirectional, interactive process~\cite{inkpen2023advancing}. However, many of these systems do not explicitly model uncertainty and may plan under unreliable assumptions. Moreover, they rarely maintain an explicit model of the knowledge gap during interaction, leading to redundant, lengthy questions that inflate interaction cost and erode user trust.
In contrast, intent-inference methods such as Bayesian methods~\cite{hoffman2024inferring, alali2024deep}, Partially Observable Markov Decision Process (POMDP) models~\cite{hadfield2016cooperative, fang2023implementing}, and Cooperative Inverse Reinforcement Learning (CIRL)~\cite{hadfield2016cooperative, tangmalinzero}, 
fail to actively reason about the intent evolution of humans over the long-horizon in order to predict future human intent. Recent LLM-based intent reasoning~\cite{liu2025idagc} offers greater expressiveness, yet its high computational cost and latency often limit deployment in real-time collaboration.
There exists an important gap in creating autonomous agents that can actively model their own knowledge gaps, reason about whether explicit dialogue or silent observation closes them at the lowest cost, and query or elicit responses accordingly.


To address these limitations, we propose an end-to-end unified human-robot joint planning system that connects the perception model and low-level drone controller with a core planning engine, supporting two complementary planning modes that operationalize the two ends of the communication spectrum: \emph{uncertainty-mitigation joint planning}, which resolves decision-critical uncertainty through explicit clarification dialogue, and \emph{real-time intent-aware collaboration}, which coordinates through implicit, nonverbal intent cues, as shown in Figure \ref{fig:overview}.
The system is mainly designed for search-and-rescue scenarios involving drones. It starts with information gathering from the surrounding environments via perception models. Next, a central interface is presented for human interactions. Depending on the task type, one of the planning modules will be activated from the core planning engine, which analyzes the information, resolves the uncertainties, and then drafts refined plans. These plans will finally be sent to the low-level controller for drone deployment.

Within the first planning mode, the robot receives a natural-language task instruction; it then categorizes planning uncertainty into structured target ambiguity and obstacle traversability, utilizing an LLM to resolve them in two stages. To eliminate target ambiguity, the agent grounds language descriptions to detected objects and iteratively refines the target candidates using tool calls and human queries. For obstacles, to minimize the expected interaction costs, we formulate hypothesis-augmented path planning and build a decision tree over traversability propositions. We then obtain an optimal querying policy via dynamic programming that guides the two-way conversation until the definitive safe path is found, so the robot asks only the questions whose answers actually change the plan.
In the second mode, the robot performs real-time cooperative planning by maintaining a probabilistic belief over the human’s latent task intent, updated online from distance and motion-direction cues together with task completion status. Conditioned on this belief, the robot selects task-level actions using coordination-aware rules that prioritize cooperative tasks and avoid redundant effort on independent tasks. 
Together, the two planning modes enable a practical human-robot teammate that talks when talking matters and stays silent when observation suffices. The former resolves decision-critical uncertainty at minimal interaction cost, while the latter adapts online to evolving human intent and priorities in open-world collaboration. They also lay the groundwork for agents that move fluidly along this communication spectrum, escalating from implicit inference to explicit clarification as behavioral confidence drops.

For the experiments, we design two tasks, human-instructed planning under uncertainty and human-participated cooperation without communication, to evaluate the two modules. The core planning engine is first evaluated in Gazebo simulation, then integrated with a perception model, a voice interface, and a low-level drone controller for real-world deployment. Compared with the baselines, the uncertainty-mitigation module reduces the interaction cost by $51.9\%$ while maintaining a $100\%$ success rate, and the intent-aware module achieves $74.3\%$ average target prediction, reducing the total task execution time by $25.4\%$.

The main contributions of this paper are as follows:
\begin{itemize}
    
    \item We propose an uncertainty-mitigation joint planning module that resolves decision-critical uncertainty through bidirectional clarification dialogue. The module localizes both target ambiguity and traversability uncertainty, and derives a cost-optimal querying strategy that steers the conversation toward a valid and reliable plan with minimal interaction cost.

    \item We propose a real-time intent-aware collaboration module that reads implicit, nonverbal intent cues, maintaining an online probabilistic belief over human intent and task progress so the robot can adapt its cooperative strategy in real time without explicit communication or retraining.

    \item We design and prototype a human-UAV collaborative planning system that integrates the two planning modes with a voice dialogue interface, perception models, and low-level controllers. Experiments in both Gazebo simulation and open-world environments verify its effectiveness.
\end{itemize}

%% file: methodology.tex
\section{Core Planning Engine}
\label{chapter:Method}




The core planning engine, as the most critical component of the entire system, distinguishes this approach from other solutions. It processes information provided by the perception model, resolves uncertainties within it, and ultimately generates high-level instructions as output. The uncertainties in the two planning modes are handled in distinct ways, which will be detailed separately in this chapter.

\subsection{Uncertainty-Mitigation Joint Planning}

We formulate the uncertainty-mitigation joint planning problem as follows. Consider an agent equipped with a perception module. At the onset of each task, a natural language instruction $I$ is received, which may be ambiguous. The perception module then constructs a structured semantic map $\mathcal{M}$ containing detected objects $O = \{o_1, o_2, \cdots, o_n\}$, their positions, and extracted attributes such as color and obstacle probability. The expected output is a planned path consisting of a sequence of way-points. 
    

In this task, uncertainty originates from semantic ambiguity in human commands, and perceptual uncertainty stems from partial observability. From the perspective of drafting a plan, these uncertainties mainly impact two factors: (1) \textit{target selection} and (2) \textit{object traversability}. Target uncertainty inevitably alters the entire plan, as the destination point varies. In contrast, the impact of the traversability of each object is conditional and can only be evaluated once a temporary target is established: several conflicting hypotheses regarding traversability paths can only be generated based on that for comparison. Therefore, the two factors should be analyzed separately and in a specific order. Note that, to reduce complexity, the uncertainty impact on obstacle-affected ranges is simply handled by introducing safety margins and is thus not considered explicitly. 

To handle the uncertainties, we propose an uncertainty-mitigation joint planning module with human-robot two-way communication, as shown in Figure \ref{fig:overview}. First, the LLM drafts a plan prototype based on $I$, $\mathcal{M}$, and the current knowledge base $\mathcal{K}$. Each step is selected from an available action list. If the action is navigation, a target description in natural language will be added. 
Second, the agent grounds these natural language descriptions into the physical properties of detected objects within the operational field, utilizing both $\mathcal{M}$ and $\mathcal{K}$. 
This process begins with the LLM assigning a prior probability to each object, indicating its likelihood of being the intended target. 
Thus, uncertainty in target selection occurs when the set of candidate objects $\mathcal{C}_{target} = \{ o_i \in O \mid P_{LLM}(o_i = \textit{target} | I, \mathcal{M}, \mathcal{K}) \geq \tau \}$ contains multiple elements, where the hyper-parameter $\tau$ is a threshold.
In such cases, the agent utilizes available tools to acquire additional information, such as size/distance comparisons, random selection, or querying the human--and subsequently updates $\mathcal{K}$. This iterative process continues until a unique target is identified for each subtask.

Next, the agent addresses obstacle uncertainty by constructing a decision tree using the A* navigation algorithm with a hypothesis-augmented state space. When encountering an uncertain obstacle $o_i$ with position $v$, the algorithm adds node $u = \langle v, H \cup \{o_i\} \rangle$ to the prioritized queue, where $H$ is the current set of uncertain objects assumed to be passable, starting from $\varnothing$. Each unique $u$ will only be added once. If $\langle goal, H \rangle$ is reached for any $H$, then the corresponding path and $H$ will be recorded as part of the output, and all nodes in the queue with $H$ being a subset will be removed. Thus, the algorithm guarantees that paths with lower costs and stricter hypotheses always come first. After the algorithm is finished, a decision tree can thus be established in the order of path costs, with hypotheses being the branch conditions.

Finally, we formulate the problem as finding an optimal querying strategy with minimal total costs for querying and verification to determine the true path. Specifically, let $n$ denote the total number of objects involved in hypotheses and impacting path-planning through the generated decision tree, each assigned a prior probability $p_i \in (0, 1)$ of being passable by the perception module. The generated decision tree inherently provides a deterministic mapping $T$ from the $2^n$ possible environment configurations (the truth table) to $m$ candidate paths. To resolve the remaining uncertainties, the agent actively queries humans with a subset of unknown objects $U$ to get their traversability, incurring an interaction cost $\lambda_1$ and a verification cost $\lambda_2$ per object evaluated. 
Given that $n$ is generally small, we compute an exact dynamic programming solution over the belief state space $s \in \{0, 1, \text{unknown}\}^n$. By recursively minimizing the Bellman equation:
\begin{align}
    V(s) & = \min_{U} \left[ \lambda_1 + |U|\lambda_2 + \mathbb{E}_{b(U)}[V(s \cup b(U))] \right],\\
    V(s_0) & = 0 \quad \text{for any } s_0 \textit{ s.t.} \ |\{T(s_0)\}| = 1.
\end{align}

Here, $\{T(s_0)\}$ indicates the set of all possible paths matching belief $s_0$, while $b(U)$ represents the binary beliefs about the traversability of $U$ retrieved from human answers, with a prior possibility of $P(b(o_i))=p_i$. The framework derives an optimal querying policy $\pi$, which guarantees that the robot identifies the definitive safe path with the absolute minimum expected interaction cost. The details about the search and querying are presented in Algorithm \ref{alg:joint_planning}. 

For the example in Figure \ref{fig:overview}, the instruction ``Pick up the medicine and deliver it to the injured person'' omits the location of the medicine. The agent constructs an initial probability distribution over detected objects and refines this target belief through LLM-guided queries. Once the target is resolved, the system generates candidate paths and a decision tree conditioned on the traversability hypotheses of the fire, net, and smoke, while the yellow box is implicitly pruned as it intersects no optimal path candidate. Executing the optimal querying strategy, the agent asks the operator about the fire and smoke while deferring questions about the net: although the fire and the net equally impact path selection, the fire's high prior probability of being an obstacle drastically reduces the expected need for subsequent queries about the net, yielding a lower expected total cost.


\begin{algorithm}[htbp]
\small
\caption{\small Uncertainty-Mitigation Joint Planning}
\label{alg:joint_planning}
\textbf{Inputs:} Start Position $v_{start}$, Goal $v_{goal}$, Uncertain object list $O$ with priors $\{p_i\}_{i=1}^n$, Query cost $\lambda_1$, Verification cost $\lambda_2$ \\
\textbf{Outputs:} The definitive safe path $P^*$
\begin{algorithmic}[1]
\State Initialize priority queue $Q \gets \{\langle v_{start}, \emptyset \rangle\}$, $\mathcal{P} \gets \emptyset$
\While{$Q$ is not empty \textbf{and} $\langle v_{goal}, \emptyset \rangle$ is not reached}
    \State Pop $u = \langle v, H \rangle$ from $Q$ with the lowest path cost
    \If{$v == v_{goal}$}
        \State Add corresponding path and hypothesis $H$ to $\mathcal{P}$
        \State Remove all nodes $\langle \cdot, H' \rangle$ from $PQ$ where $H \subseteq H'$
        \State \textbf{continue}
    \EndIf
    \For{each valid neighbor $v'$ of $v$}
        \State $H' \gets H \cup \{v'\}$ \textbf{if} $v' \in O_{unc}$ \textbf{else} $H$
        \State Add unique node $u' = \langle v', H' \rangle$ to $PQ$
    \EndFor
\EndWhile
\State Extract the decision tree as a mapping $T: \{0,1\}^n \to \mathcal{P}$ evaluating configurations against collected $H$
\State Initialize memoization table $V_{memo} \gets \emptyset$, Policy $\pi^* \gets \emptyset$
\Function{MinCost}{$s$} \Comment{$s \in \{0, 1, \text{unknown}\}^n$}
\If{$|\{T(s)\}| == 1$} \Return $0$
    \EndIf
    \If{$s \in V_{memo}$} \Return $V_{memo}[s]$ 
    \EndIf
    \State $v_{min} \gets \infty$, $U^* \gets \emptyset$
    \For{each subset $U \subseteq \text{unknown}(s), U \neq \emptyset$}
        \State $cost \gets \lambda_1 + |U|\lambda_2 + \mathbb{E}_{b(U)|s}[\Call{MinCost}{s \cup b(U)}]$
        \If{$cost < v_{min}$}
            \State $v_{min} \gets cost$, $U^* \gets U$
        \EndIf
    \EndFor
    \State $V_{memo}[s] \gets v_{min}$, $\pi^*[s] \gets U^*$
    \State \Return $v_{min}$
\EndFunction
\Statex \vspace{-0.2cm}
\State $s_0 \gets (\text{unknown}, \dots, \text{unknown})$
\State \Call{ComputeMinCost}{$s_0$}
\State Query human for $\pi^*[s]$, update belief $s$, until $|\{T(s)\}| == 1$
\State \Return $T(s)$
\end{algorithmic}
\end{algorithm}

\subsection{Real-time Intent-Aware Collaboration}
\label{sec:intent_coordination}

We consider cooperative human--robot task execution in a shared workspace where a human-controlled agent and an autonomous robot operate concurrently to complete a finite set of perception-identified tasks. The human pursues an internal and unobserved objective, while the robot must adapt online using only the human's recent motion and the evolving task completion status, without explicit communication.

Unlike conventional intent inference that emphasizes goal prediction alone, our setting involves multiple task types: tasks may be independent (completable by any single agent) or cooperative (requiring simultaneous presence). Simply following the most-likely human target can therefore be suboptimal: joining the human on an independent task introduces redundant effort, while delayed convergence on a cooperative task increases synchronization costs. These asymmetric coordination costs, together with noisy motion cues and real-time control constraints, motivate a lightweight intent belief coupled with a stability-aware coordination policy.

\subsubsection{Task model and state representation}
Let $\mathcal{T}=\mathcal{T}_{\mathrm{ind}} \cup \mathcal{T}_{\mathrm{coop}}$ denote the task set, where $\mathcal{T}_{\mathrm{ind}}$ are independent tasks and $\mathcal{T}_{\mathrm{coop}}$ are cooperative tasks requiring both agents to participate. Each task $t_i$ has a known location $x_i \in \mathbb{R}^2$. At time $t$, the state is $s_t=(x_t^H,x_t^R,C_t)$, where $x_t^H$ and $x_t^R$ are the human and robot positions, and $C_t$ is the set of completed tasks. The remaining tasks are $\mathcal{T}_t=\mathcal{T}\setminus C_t$. The robot observes $x_t^H$ at each time and maintains $C_t$ online.

\subsubsection{Lightweight intent belief update}
At each time step, the robot maintains a normalized belief over remaining tasks,
\begin{equation}
p_t(i) \triangleq P(g_t=t_i \mid x_{0:t}^H, C_t), \qquad t_i \in \mathcal{T}_t,
\end{equation}
where $g_t$ denotes the human's current task target. Completed tasks are masked by enforcing $p_t(i)=0$ for $t_i\notin\mathcal{T}_t$ and renormalizing over $\mathcal{T}_t$.

We compute intent evidence from two local geometric cues: (1) distance-to-task and (2) heading alignment. For each candidate $t_i\in\mathcal{T}_t$, let $d_{t,i}=\|x_t^H-x_i\|_2$. Let the finite-difference velocity be $v_t=x_t^H-x_{t-1}^H$. When $\|v_t\|_2$ is below a threshold $\varepsilon$, we treat the heading cue as neutral; otherwise we compute alignment as the cosine between the human heading and the direction to the task,
$c_{t,i}=\hat{v}_t^\top \hat{u}_{t,i}$, where $\hat{v}_t=v_t/\|v_t\|_2$ and $\hat{u}_{t,i}=(x_i-x_t^H)/\max(\|x_i-x_t^H\|_2,\varepsilon)$.

We convert these cues into a soft evidence score
\begin{equation}
\tilde{p}_t(i)=\exp\!\big(-\alpha\, d_{t,i} + \beta\, c_{t,i}\big),
\end{equation}
with $\alpha>0$ controlling distance sensitivity and $\beta\ge 0$ controlling directional influence. To accumulate evidence while remaining responsive, we apply exponential smoothing and then normalize:
\begin{equation}
p_t(i) \propto (1-\gamma)\,p_{t-1}(i) + \gamma\,\tilde{p}_t(i), \qquad \gamma\in(0,1].
\end{equation}
This update is $O(|\mathcal{T}_t|)$ per step and runs at control rate.

\subsubsection{Coordination-aware task selection}
Given $p_t$, the robot selects a task-level target $a_t\in\mathcal{T}_t$ and executes motion via a low-level controller. Let $\hat{g}_t=\arg\max_{t_i\in\mathcal{T}_t} p_t(i)$ denote the most likely human target, with confidence $\rho_t=p_t(\hat{g}_t)$. The robot switches its target only when $\rho_t \ge \tau_{\mathrm{intent}}$ and it is not already committed, where commitment is defined by being within a radius $r_{\mathrm{commit}}$ of the current target, as long as that target has not yet been completed. If $\rho_t < \tau_{\mathrm{intent}}$, the robot maintains its current target; if no target is assigned, it selects the nearest remaining task. This prevents oscillatory switching under ambiguous intent estimates.

When switching is allowed, the action depends on the task type. If $\hat{g}_t\in\mathcal{T}_{\mathrm{coop}}$, the robot sets $a_t=\hat{g}_t$ to reduce synchronization delay; if it arrives first, it waits up to a bounded timeout before re-evaluating the belief. If instead $\hat{g}_t\in\mathcal{T}_{\mathrm{ind}}$, the robot selects complementary work by choosing the nearest remaining independent task other than $\hat{g}_t$ (when available), thereby avoiding redundant effort; if no such task exists, it selects $\hat{g}_t$.

By coupling a lightweight geometric intent belief with task-type-aware role allocation and explicit stability gating, the robot adapts online to evolving human behavior, thereby minimizing redundant work, reducing cooperative waiting time, and maintaining stable real-time execution.

%% file: experiments.tex
\section{Details of System Design}


To deploy and evaluate the proposed core planning engine in the real world, we established an end-to-end system connecting it with perception and low-level robot execution, which
provides a central interface that acts as a voice assistant. By interpreting the operator's voice instructions, it generates actionable scripts to carry out the entire task. 

For environmental perception, a Vision-Language Model (VLM) processes RGB-D sensor data to build a 3D semantic representation, which is presented in the central interface for real-time situational awareness.
Based on the task setting, one planning module is then activated.
In uncertainty-mitigation joint planning, when querying is required, the agent pauses execution and elicits a question in natural language, which the voice module speaks aloud to ask the operator for clarification; once the uncertainty is removed, the module generates a planned path in precise waypoints.
In intent-aware collaboration, the system monitors the real-time positions of the human and the robot, infers human intent automatically, and issues high-level control commands.
These outputs are transmitted to the low-level drone controller, pre-trained with reinforcement learning following \cite{yu2025equivariant}, to actuate the UAV and carry out the mission. The perception and voice modules are implemented as follows:

\begin{figure}[t]
    \centering
    \vspace{0.2cm}
    \begin{subfigure}[b]{0.2\textwidth}
        \centering
        \includegraphics[width=\textwidth]{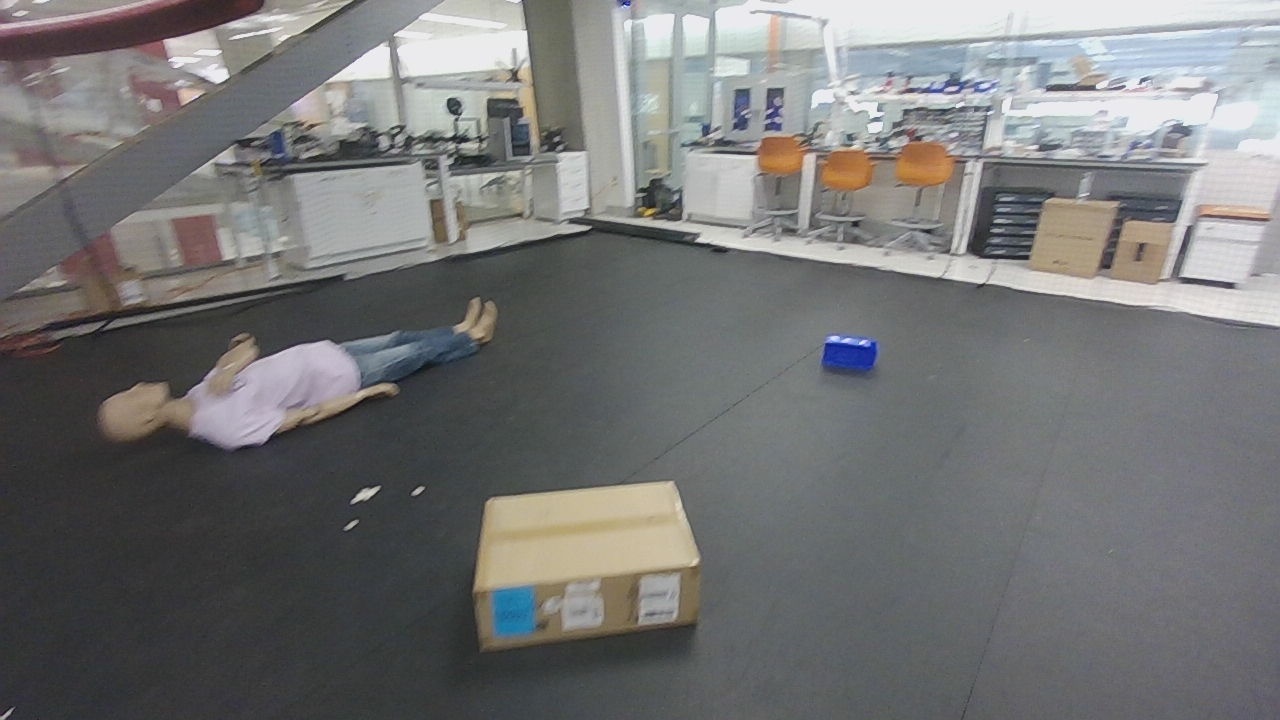}
    \end{subfigure}\hspace{0.5mm}
    \begin{subfigure}[b]{0.2\textwidth}
        \centering
        \includegraphics[width=\textwidth]{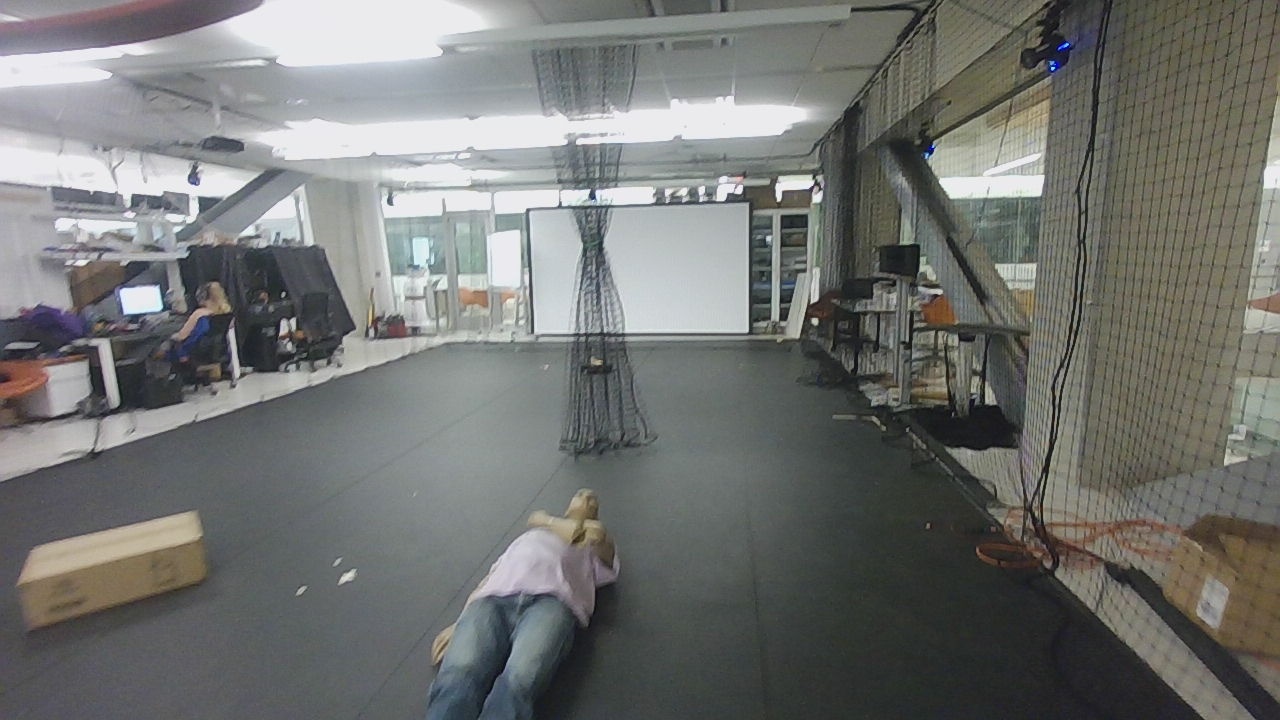}
    \end{subfigure}\hspace{0.5mm}%
    
    \nointerlineskip 
    \vspace{0.5mm}   

    \begin{subfigure}[b]{0.2\textwidth}
        \centering
        \includegraphics[width=\textwidth]{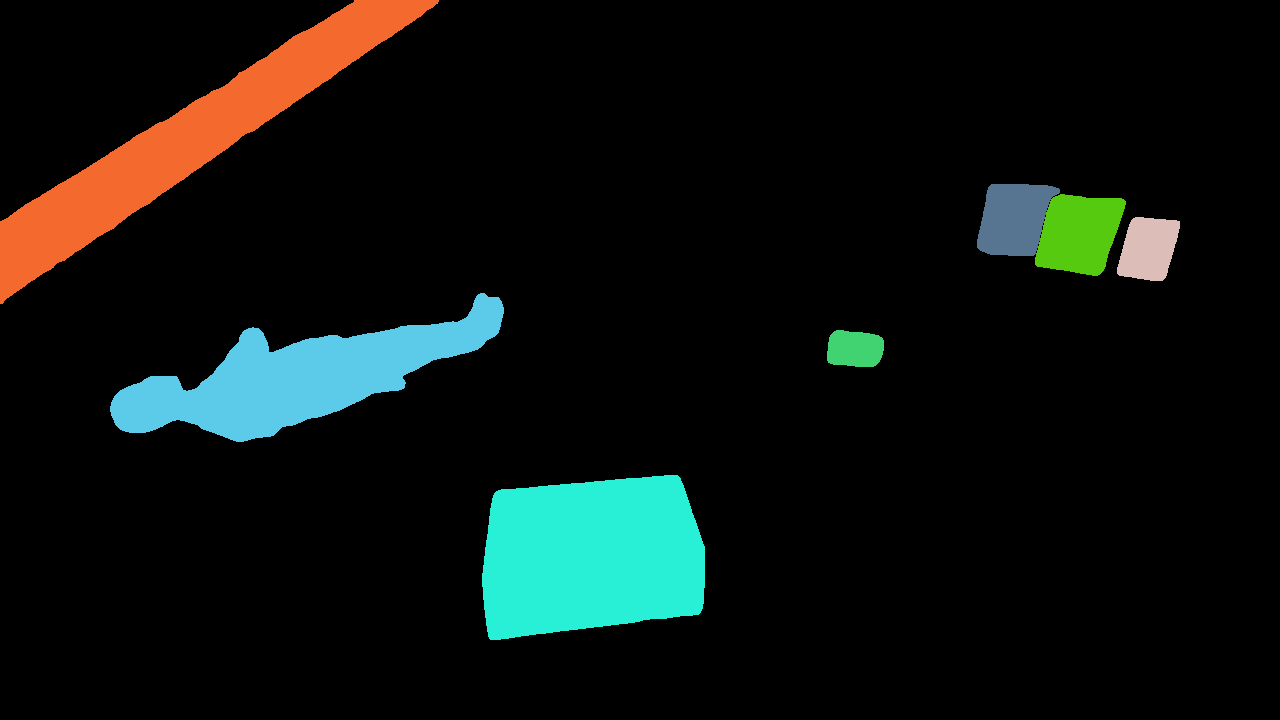}
    \end{subfigure}\hspace{0.5mm}%
    \begin{subfigure}[b]{0.2\textwidth}
        \centering
        \includegraphics[width=\textwidth]{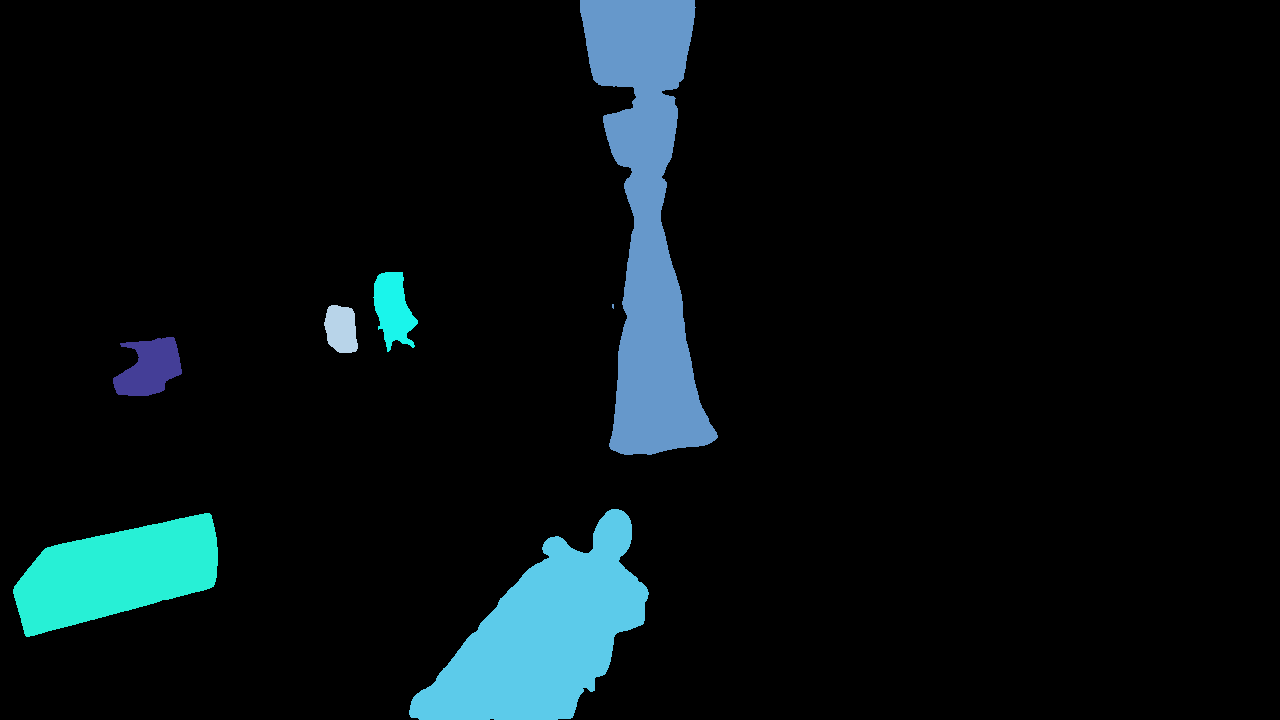}
    \end{subfigure}\hspace{0.5mm}%

    \caption{Semantic Fusion: The top row displays the original RGB inputs. The bottom row illustrates the semantic segmentation masks after the recursive fusion strategy.}
    \label{fig:temporal_fusion_grid}
    \vspace{-0.5cm}
\end{figure}

\begin{figure}
    \centering
    \begin{subfigure}[b]{0.4\textwidth}
        \centering
        \includegraphics[width=\textwidth]{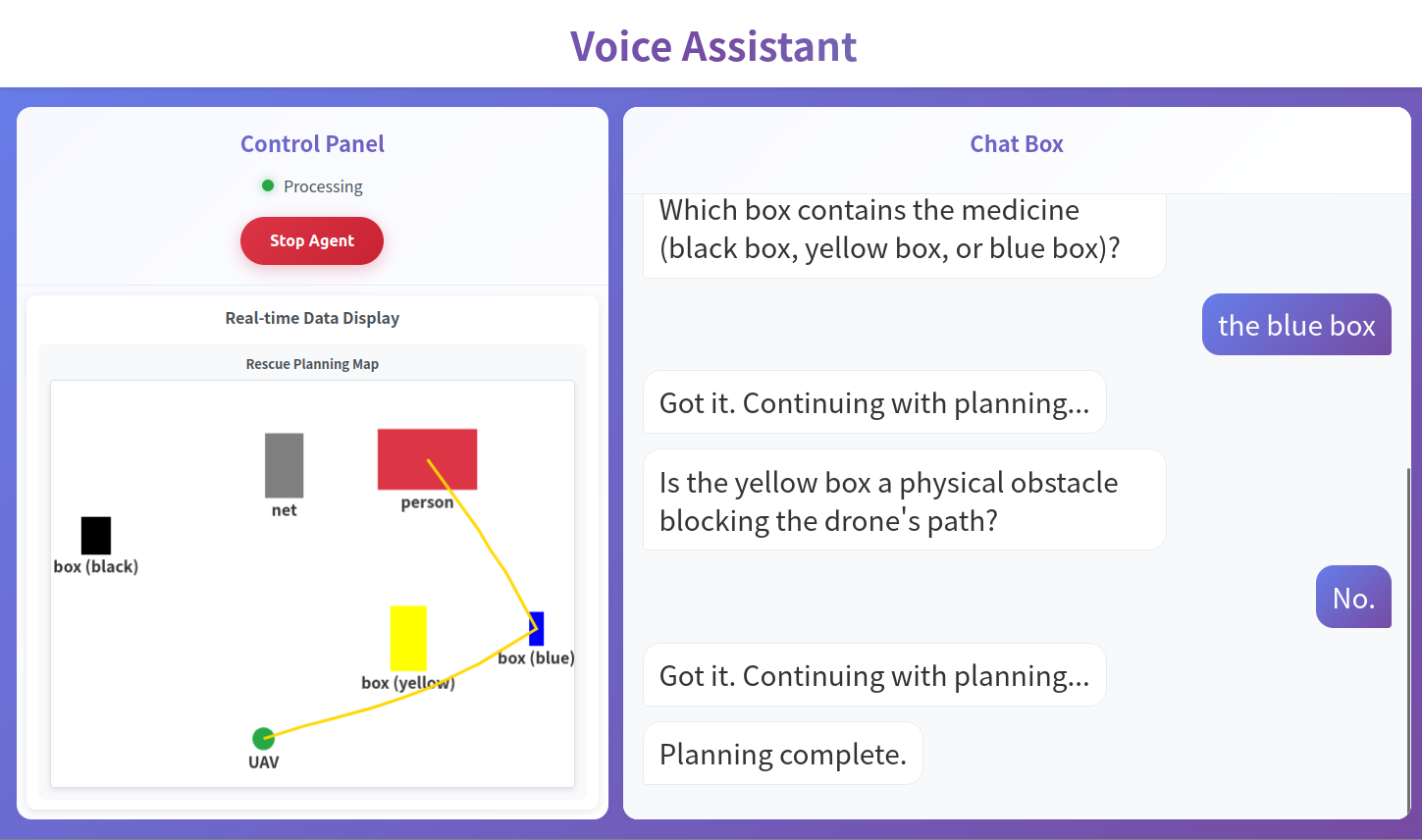}
    \end{subfigure}
    
    \vspace{0.5mm}
    
    \begin{subfigure}[b]{0.4\textwidth}
        \centering
        \includegraphics[width=\textwidth]{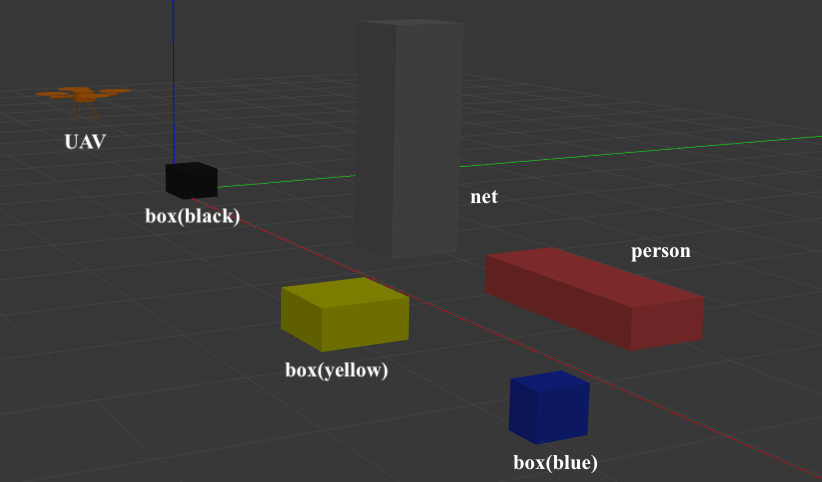}
    \end{subfigure}\hspace{0.5mm}%
    \caption{Voice interface and Gazebo simulation environment.}
    \vspace{-0.6cm}
    \label{fig:voice}
\end{figure}

\subsubsection{VLM-Based 2D Perception}
A fundamental limitation of conventional perception systems is their reliance on fixed-label classifiers, which restrict adaptability in open-world navigation. 
To overcome this, our system employs the Grounded-SAM framework~\cite{ren2024grounded} to enable arbitrary natural language interaction, allowing the UAV to retrieve target objects by comparing prompt embeddings against stored 3D language-aligned descriptors. Specifically in this framework, Grounding DINO~\cite{liu2024grounding} detects objects and produces 2D bounding boxes, which are subsequently refined into high-quality segmentation masks using SAM~\cite{kirillov2023segment}.

\subsubsection{3D Semantic Map Representation}
We reconstruct the environment using 3D Gaussian Splatting~\cite{kerbl20233d}, augmenting each anisotropic Gaussian with a semantic label ID, a confidence score $c \in [0, 1]$, and feature embeddings. This representation supports efficient rasterization-based rendering, reprojecting the 3D semantic state onto the image plane to provide geometric and semantic references for incoming 2D observations.
To ensure consistent data association, we implement a two-stage verification process. The system first evaluates cosine similarity between incoming 2D image features and latent 3D embeddings. Candidate matches are then validated against rendered semantic reprojections, providing a direct geometric reference. This hierarchical alignment fuses new observations into the correct spatial clusters, maintaining a stable 3D representation.

\subsubsection{Recursive Semantic Fusion}
Following association, a recursive fusion strategy maintains consistency across sequential frames. Although SAM provides high-quality masks, inaccurate boundaries in 2D segmentation models often corrupt 3D object geometry. We mitigate this by evaluating label consistency solely for pixels satisfying a strict depth continuity threshold ($\Delta z < 0.05$m). Matching labels trigger a confidence update via weighted averaging, whereas conflicts reduce the existing Gaussian confidence by a decay factor ($\gamma = 0.85$). To further harden the map against edge noise, confidence updates are weighted by a Euclidean distance transform of the 2D mask, prioritizing central pixels over noisy boundaries. This allows the 3D representation to self-correct over time, as shown in Fig.~\ref{fig:temporal_fusion_grid}.

\subsubsection{Object Localization and Outlier Mitigation}
Segmented 3D object candidates are converted into axis-aligned bounding boxes in the navigation space. To remove sparse reconstruction outliers, we discard extreme points along each principal axis and retain the central 92\% of the cluster when computing object bounds. Only candidates within the arena boundaries are retained for planning. By maintaining the DINO-based feature embeddings within the 3D map, our system supports natural language interaction. Users can provide a prompt, which is embedded into the same latent space and compared against the stored 3D object features to retrieve the corresponding target object for planning.

\subsubsection{Voice Module}
The voice system, implemented as a web application, provides hands-free interaction using OpenAI Whisper for speech-to-text transcription, GPT-4o for dialogue understanding and command interpretation, and the OpenAI TTS API for speech synthesis. 
Integrated with the planning module, it handles multi-turn communication and planning; the voice agent interface is shown in Figure \ref{fig:voice}.

\section{Experimental Results}

\subsection{Simulation Verification}






To evaluate the perception models exclusively, we first conduct simulation experiments in Gazebo, which is fully integrated with ROS2, as shown in Fig. 3. 



\subsubsection{Evaluation of the Uncertainty-Mitigation Joint Planning.} 

We compare the proposed uncertainty-mitigation module with two baselines: no querying, which plans directly without communication or uncertainty resolution, and exhaustive querying, which queries the human about every uncertainty.

We evaluate these methods in Gazebo on 20 simple scenarios with a single type of uncertainty and 5 complex scenarios with multiple sources. In each case, uncertainty arises from multiple target candidates or undetermined obstacles, and the agent needs additional information (e.g., target distance, size, color, or obstacle traversability) to provide the optimal plan. Each case is run 5 times to mitigate the randomness introduced by the LLM. We use ChatGPT-5.2 as the backbone with structured output and chain of thought disabled, and set the query cost $\lambda_1=10$ and verification cost $\lambda_2=1$.

The results are illustrated in Figure \ref{fig:simulation_result_zeyu}. The exhaustive baseline and the proposed method achieved a $100\%$ success rate in both simple and complex scenarios, while the no-query baseline only achieved $71\%$ in simple tasks and $40\%$ in complex tasks due to a lack of information exchange with humans. Typical failures occur when the agent misidentifies the target or makes false assumptions given a vague command; even in successful cases, the no-query baseline plans conservatively to avoid all possible uncertainties and thus generates longer paths. On the other hand, the proposed method significantly reduces the query times by $56.8\%$ and the total token usage by $30.3\%$ with the adoption of hypothesis-based A* and dynamic programming to obtain the optimal query strategy.

\subsubsection{Evaluation of Real-Time Intent-Aware Collaboration}

We evaluate the intent-aware module in simulation against a non-cooperative baseline, where the robot selects the nearest uncompleted task as its target without modeling human intent or performing task-level coordination.

Four layouts are considered, each with two independent tasks and one cooperative task. Three are synthetic with distinct spatial configurations, and one is derived from the real-world setup with denser placement. For each layout, both methods are tested under three human behavior cases, including two rational and one ambiguous pattern, across two human operators, yielding 24 runs per method.

The intent recognition parameters are fixed at $\alpha = 0.3$ and $\beta = 1.0$ with a confidence threshold of $\tau_{\mathrm{intent}} = 0.5$. The commitment distance $r_{\mathrm{commit}}$ is set to 1.5 m for synthetic layouts and 1.0 m for the denser layout. If the robot reaches a cooperative task without the human, it waits up to 5 s before re-planning; the human follows a similar strategy. Each episode terminates upon task completion, and results are reported as mean $\pm$ standard error.

    
    

\begin{figure}[t]
    \centering
    \vspace{0.2cm}
    \begin{subfigure}{0.49\columnwidth}
        \centering
        \includegraphics[width=\linewidth]{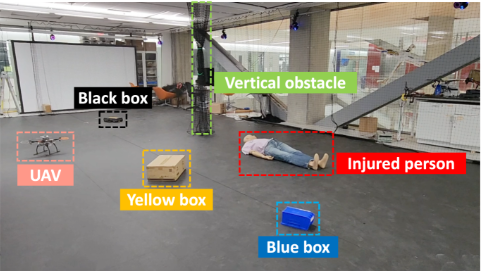}
        \caption{Planning Mode 1}
        \label{fig:joint_planning}
    \end{subfigure}
    \hspace{0.5mm}%
    \begin{subfigure}{0.49\columnwidth}
        \centering
        \includegraphics[width=\linewidth]{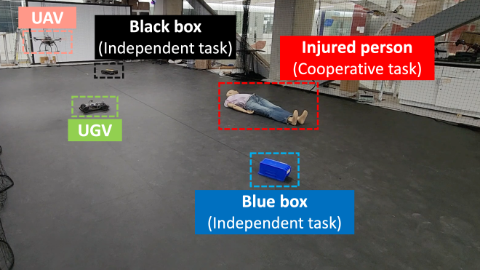}
        \caption{Planning Mode 2}
        \label{fig:intent_aware}
    \end{subfigure}
    
    \caption{Evaluation setups: (a) shows the joint planning configuration, while (b) details the intent-aware setup.}
    \label{fig:experimental_setup}
    \vspace{-0.6cm}
\end{figure}

\begin{table}[ht]
\centering
\caption{Simulation performance of the intent-aware cooperation module compared to
the non-cooperative baseline.}
\label{tab:overall-simulation}
\resizebox{\linewidth}{!}{
\begin{tabular}{lcccc}
\toprule
\textbf{Method} & \textbf{Time (s)} & \textbf{Total Dist (m)} & \textbf{Human Dist (m)} \\
\midrule
Proposed    & 14.83 $\pm$ 0.79 & 24.46 $\pm$ 1.41 & 12.26 $\pm$ 0.76 \\
Non-Cooperative & 19.26 $\pm$ 0.90 & 27.40 $\pm$ 1.33 & 14.90 $\pm$ 0.90 \\
\bottomrule
\end{tabular}
}
\vspace{-0.2cm}
\end{table}


Table~\ref{tab:overall-simulation} summarizes overall simulation performance, where travel distance also serves as a proxy for motion-related energy cost. The proposed method reduces execution time by 23.0\%, total travel distance by 10.7\%, and human travel distance by 17.7\% compared to the non-cooperative baseline. The time reduction reflects fewer coordination delays, while the larger decrease in human distance indicates that the drone assumes more of the workload.

\begin{table}[ht]
\centering
\caption{Intent recognition performance measured by average true-target probability and top-1 accuracy.}
\label{tab:avg-intent}
\resizebox{\linewidth}{!}{
\begin{tabular}{lccc}
\toprule
 \textbf{Metric} & \textbf{Rational} & \textbf{Ambiguous} & \textbf{Overall} \\
\midrule
Avg. True-Target Prob. & 75.2\% $\pm$ 2.2\% & 65.3\% $\pm$ 3.0\% & 71.9\% $\pm$ 2.0\% \\
Top-1 Acc. & 90.6\% $\pm$ 2.5\% & 70.7\% $\pm$ 3.2\% & 84.0\% $\pm$ 2.8\% \\
\bottomrule
\end{tabular}}
\end{table}

Table~\ref{tab:avg-intent} shows the intent recognition performance, where the true target at each step is approximated by the next task completed by the human. The overall true-target probability reaches 71.9\%, well above the 33\% uniform baseline, and the top-1 accuracy of 84.0\% means the highest-belief task matches the subsequently completed task in most cases. Performance is slightly lower in ambiguous scenarios, but the gap remains modest.

\begin{figure}[ht]
    \centering
    \includegraphics[width=0.82\linewidth]{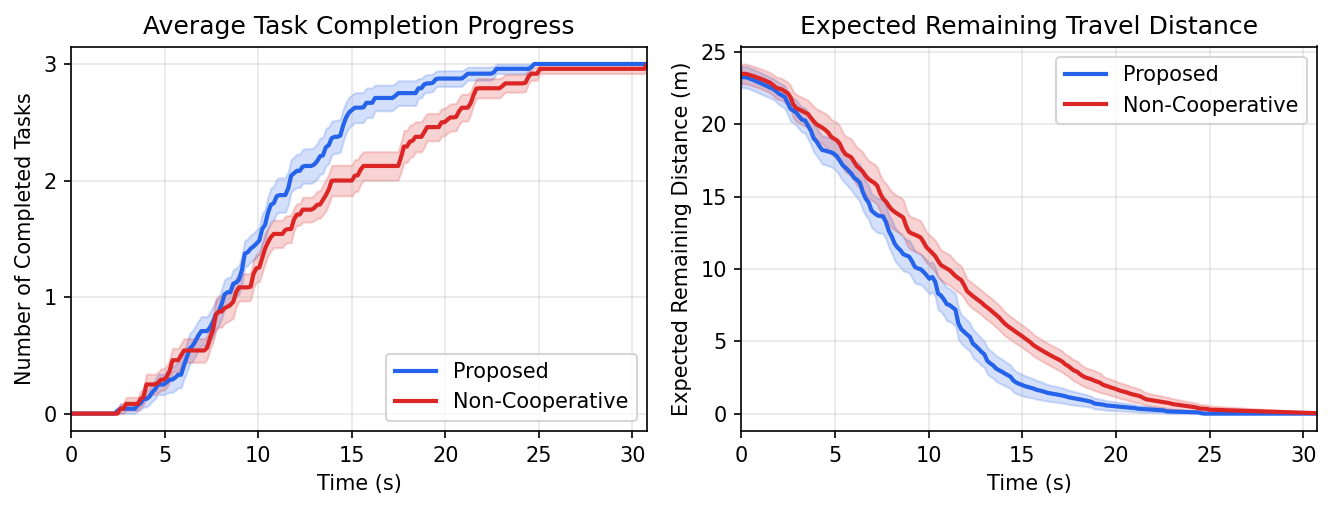}
    \caption{Task completion progress (left) and expected remaining travel distance (right) over time for the proposed method and non-cooperative baseline.}
    \label{fig:yuxin-experiment-comparison}
    \vspace{-0.2cm}
\end{figure}

Figure~\ref{fig:yuxin-experiment-comparison} illustrates the temporal evolution of both methods. In the left panel, performance is similar early on, as the non-cooperative baseline completes nearby independent tasks efficiently. However, the proposed method advances faster during the cooperative phase and finishes earlier overall. The right panel shows the expected remaining travel distance based on shortest-path estimates; the steeper decline under the proposed method reflects more efficient coordination throughout execution.

\subsection{Real-world Deployment}

\subsubsection{Experimental Setup and Scene Composition}

As shown in Fig.~\ref{fig:experimental_setup}, experiments were conducted in a $12\,\text{m} \times 6\,\text{m}$ motion-captured indoor space configured for two distinct evaluation scenarios.
For the human-robot joint planning scenario, the scene included five entities: a human mannequin (injured person), a black box, and a blue box acting as potential task targets, alongside a yellow box and a vertical net serving as static obstacles. 
For the intent-aware planning scenario, the environment featured two independent tasks (the black and blue boxes) and one cooperative task (the injured person) shared between the autonomous UAV and a human-teleoperated UGV.
Both high-level planning modules operated at 100 Hz to ensure real-time coordination.

To construct the semantic map, RGB-D images were acquired using an ORBBEC Gemini 336L camera, while a Vicon motion-capture system provided ground-truth camera poses at 200 Hz. This setup enabled the real-time fusion of observations into a consistent global 3D frame. The resulting refined 2D bounding box coordinates and their corresponding Grounding DINO confidence scores for the objects are summarized in Table~\ref{tab:object-detection}.

\begin{table}[ht]
\centering
\caption{2D Object Localizations and Confidence Scores.}
\label{tab:object-detection}
\resizebox{0.5\textwidth}{!}{
\begin{tabular}{lccccc}
\toprule
\textbf{Attribute} & \textbf{Blue box} & \textbf{Black box} & \textbf{People} & \textbf{Yellow Box} & \textbf{Net} \\
\midrule
\textbf{Confidence} & 0.8147 & 0.6342 & 0.5336 & 0.4644 & 0.3371 \\
\addlinespace
\textbf{Range} $x$& $[7.39, 7.64]$ & $[-0.02, 0.47]$ & $[4.89, 6.53]$ & $[5.10, 5.70]$ & $[3.02, 3.66]$ \\
\textbf{Range} $y$ & $[-1.04, -0.68]$ & $[-0.09, 0.30]$ & $[0.58, 1.21]$ & $[-1.30, -0.62]$ & $[0.49, 1.17]$ \\
\bottomrule
\end{tabular}
}
\end{table}

To physically execute these planned behaviors, a custom quadrotor platform was developed, equipped with an onboard NVIDIA Jetson TX2 flight computer and an IMU sensor.
The core flight software and Extended Kalman Filter (EKF) for state estimation were implemented in C++, communicating via ROS2 with the PyTorch-based RL models implemented in Python. 
Operating at 200Hz, this dedicated low-level RL controller seamlessly tracked and actuated the desired waypoints generated by the high-level planning modules.

\subsubsection{Evaluation of Uncertainty-mitigation Joint Planning}

We first evaluate the joint planning model using the same search-and-rescue scenario as in the simulation. We run three times with identical object locations, while a different command is given in each task (``Get the medicine from the box and deliver it to the person.''; ``Pick up the medicine from the box, deliver it to the person and back to your initial position.''; ``Pick up medicine from the black box and then take the bandage from the blue box and deliver both of them to the person''). All communications are done by voice. The planning process is repeated 5 times to mitigate randomness, and one run is deployed to record the success rate. The results are shown in Table \ref{tab:real_world_zeyu}.

In the first two commands, the box is not specified; thus, the passive baseline always drives the UAV to the yellow box, which is nearest to the UAV but is supposed to be an obstacle, causing a significantly low success rate. On the other hand, the other two methods always ask a human first to find out the true box containing the medicine. However, as both the yellow box and the vertical obstacle are considered potential obstacles for the UAV, our proposed method only asks the human when any of the obstacles impact the planning, while the exhaustive baseline always verifies the two objects first and then starts to plan. As a result, our method reaches $100\%$ success rate while reducing the number of queries by $51.9\%$ and $16.4\%$ of the total tokens used. 

\begin{figure}
    \centering
    \vspace{0.2cm}
    \includegraphics[width=0.78\linewidth]{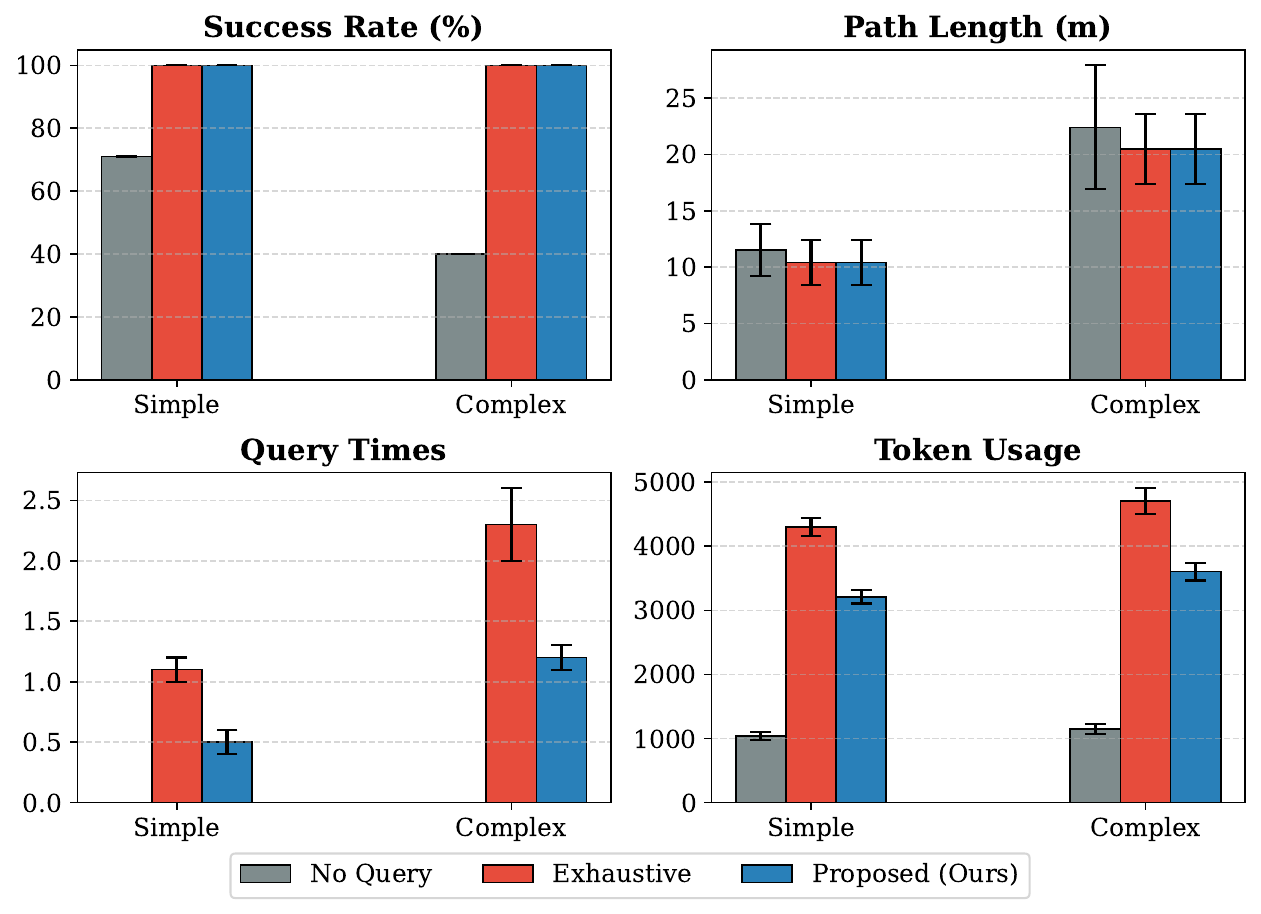}
    \caption{Visualized results of uncertainty-mitigation in Gazebo simulations. Note that we only compare paths successfully generated by all three methods for length comparison.}
    \label{fig:simulation_result_zeyu}
    \vspace{-0.6cm}
\end{figure}

\subsubsection{Evaluation of Intent-Aware Cooperation} 
We evaluate the intent-aware cooperation module on the physical system using the same two methods and human behavior cases as in simulation, with identical intent inference parameters.

Table~\ref{tab:real_world_yuxin} summarizes the results. Compared to the non-cooperative baseline, the proposed method reduces execution time by 25.4\%, total travel distance by 17.9\%, and human travel distance by 18.3\%. Execution time is naturally longer than in simulation due to physical dynamics, actuation limits, and sensing noise. Nevertheless, the relative performance improvement remains consistent, indicating effective transfer from simulation to real-world deployment.

For intent recognition, the average true-target probability reaches 74.3\%
, and the top-1 accuracy reaches 95.0\%
, both well above a uniform baseline. 
The high top-1 accuracy indicates that the inferred belief most often assigns the highest probability to the task subsequently completed by the human. Despite real-world disturbances, the belief remains aligned with observed human actions, suggesting stable intent estimation under physical execution constraints.

\begin{table}[h]
\centering
\caption{Real-world performance of the intent-aware cooperation module compared to the non-cooperative baseline.}
\label{tab:real_world_yuxin}
\resizebox{\linewidth}{!}{
\begin{tabular}{lccc}
\toprule
\textbf{Method} & \textbf{Time (s)} & \textbf{Total Dist (m)} & \textbf{Human Dist (m)}\\
\midrule
Proposed    & 20.59 $\pm$ 1.25 & 16.31 $\pm$ 1.91 & 9.22 $\pm$ 2.66 \\
Non-Cooperative & 27.61 $\pm$ 2.99 & 19.86 $\pm$ 2.88 & 11.29 $\pm$ 2.38 \\
\bottomrule
\end{tabular}
}
\vspace{-0.45cm}
\end{table}



 

 








\begin{table}[h]
\centering
\caption{Real-world experimental results for uncertainty-mitigation joint planning, averaged on three commands.}
\label{tab:real_world_zeyu}
\resizebox{0.47\textwidth}{!}{
\begin{tabular}{lcccc}
\toprule
\textbf{Method} & \textbf{Success (\%)} & \textbf{Queries} & \textbf{Tokens} & \textbf{Time (s)} \\
\midrule
Passive & $33.3$ & $0.0$ & $1079.3\pm50.2$ & $8.80\pm1.07$\\
Exhaustive & $100$ & $2.7$ & $4109.0\pm81.1$ & $13.35\pm1.51$\\
Ours & 100 & $1.3$ & $3434.8 \pm 74.2$ & $11.84\pm1.47$\\
\bottomrule
\end{tabular}}
\vspace{-0.5cm}
\end{table}